\documentclass{article}
\usepackage{graphicx,mlspconf}
\usepackage{layout}
\usepackage{bmpsize}
\usepackage{charter}
\usepackage{wrapfig}
\usepackage{xcolor}
\usepackage{amsmath}
\usepackage{amssymb}
\usepackage{amsthm}
\usepackage{lscape}
\usepackage{array}
\usepackage{enumerate}
\usepackage[nice]{nicefrac}
\usepackage{stmaryrd}
\usepackage{float}
\usepackage{caption}
\usepackage{subcaption}
\usepackage{float}
\usepackage{etoolbox}

\usepackage{algorithm}
\usepackage{algpseudocode}

\usepackage[utf8]{inputenc}
\usepackage[final]{pdfpages}
\usepackage{bbm}
\usepackage{mathtools}
\usepackage{booktabs}
\usepackage[framemethod=TikZ]{mdframed}
\usepackage{enumitem}
\usepackage{multirow}

\usepackage{hyperref}
\definecolor{linkcolor}{RGB}{128,0,128}
\definecolor{citecolor}{RGB}{83,83,182}

\hypersetup{colorlinks,citecolor={blue}, linkcolor={citecolor}}

\providetoggle{anonymous}
\settoggle{anonymous}{false}

\makeatletter
\renewcommand{\paragraph}{%
  \@startsection{paragraph}{4}%
  {\z@}{0.5ex \@plus .001ex \@minus .001ex}{-1em}%
  {\normalfont\normalsize\bfseries}%
}
\makeatother
\newcommand\vspacesize{-2mm}
\newcommand{\sizefig}{5cm}

\usepackage{array}

\def\Sigmab{\boldsymbol\Sigma}

\usepackage{tabularx}

\def\G{\pi}

\def\GG{\boldsymbol\G}

\def\abf{\mathbf{a}}
\def\bbf{\mathbf{b}}

\def\ubf{{\bf u}}

\def\X{{\bf X}}
\def\Y{{\bf Y}}

\def\R{{\mathbb{R}}}
\def\Abf{{\mathbf{A}}}

\def\Pbf{{\mathbf{P}}}

\def\Ib{{\mathbf{I}}}
\def\Mbf{{\mathbf{M}}}

\def\U{{\mathbf{U}}}

\def\Ocal{{\mathcal{O}}}

\def\tr{{\operatorname{tr}}}
\def\one{{\mathbf{1}}}

\newcommand{\xbf}{\mathbf{x}}

\newcommand{\ybf}{\mathbf{y}}

\newcommand{\ie}{\textit{i.e.}}
\newcommand{\eg}{\textit{e.g.}}

\newcommand{\D}{\Delta}

\newcommand{\sym}{\operatorname{sym}}

\def\bbf{{\mathbf{b}}}

\newcommand{\OT}{\operatorname{OT}}

\newcommand{\St}{\operatorname{St}}

\usepackage{mdframed}
\usepackage{amssymb}
\usepackage{pifont}

\newtheorem*{proposition*}{Proposition}
\newtheorem*{lemma*}{Lemma}
\newtheorem*{theorem*}{Theorem}
\newtheorem*{corollary*}{Corollary}

\newtheorem*{assumption*}{\assumptionnumber}
\providecommand{\assumptionnumber}{}


\usepackage{thmtools} 
\usepackage{thm-restate}

\declaretheoremstyle[
    headfont=\bfseries, 
    bodyfont=\normalfont\itshape,
    headpunct={},
    spacebelow=\parsep,
    spaceabove=\parsep,
    mdframed={
        innertopmargin=6pt,
        innerbottommargin=6pt, 
        skipabove=\parsep, 
        skipbelow=\parsep} 
]{framedstyle}

\declaretheorem[name=Lemma]{lemma}

\newtheorem*{prob*}{Problem}
\newtheorem*{obj*}{Objective}

\newcommand{\vertiii}[1]{{\left\vert\kern-0.25ex\left\vert\kern-0.25ex\left\vert #1 
    \right\vert\kern-0.25ex\right\vert\kern-0.25ex\right\vert}}

\usepackage{thmtools} 
\usepackage{thm-restate}

\iftoggle{anonymous}{
\copyrightnotice{U.S.\ Government work not protected by U.S.\ copyright}

\copyrightnotice{978-1-6654-8547-0/22/\$31.00 {\copyright}2023 Crown}

\copyrightnotice{978-1-6654-8547-0/22/\$31.00 {\copyright}2023 European Union}

\copyrightnotice{979-8-3503-2411-2/23/\$31.00 {\copyright}2023 IEEE}

\toappear{2023 IEEE International Workshop on Machine Learning for Signal Processing, Sept.\ 17--20, 2023, Rome, Italy}

\name{Anonymous\thanks{Anonymous.}}
\address{Anonymous}

}{\name{Antoine Collas$^1$, Titouan Vayer$^2$, Rémi Flamary$^3$, Arnaud Breloy$^4$}
\address{
$^1$Université Paris-Saclay, Inria, CEA\\
$^2$Université Lyon, Inria, CNRS, ENS Lyon, UCB Lyon 1, LIP  UMR 5668\\
$^3$Ecole Polytechnique, Institut Polytechnique de Paris, CMAP, UMR 7641\\
$^4$LEME, Université Paris Nanterre
}
}

\title{Entropic Wasserstein Component Analysis}

\begin{document}

\maketitle

\begin{abstract}
    Dimension reduction (DR) methods provide systematic approaches for analyzing high-dimensional data. A key requirement for DR is to incorporate global dependencies among original and embedded samples while preserving clusters in the embedding space. To achieve this, we 
    combine the principles of optimal transport (OT) and principal component analysis (PCA). Our method seeks the best linear subspace that minimizes reconstruction error using entropic OT, which naturally encodes the neighborhood information of the samples. From an algorithmic standpoint, we propose an efficient block-majorization-minimization solver over the Stiefel manifold. Our experimental results demonstrate that our approach can effectively preserve high-dimensional clusters, leading to more interpretable and effective embeddings.
    \iftoggle{anonymous}{}{Python code of the algorithms and experiments is available online\footnote{\url{https://github.com/antoinecollas/Entropic_Wasserstein_Component_Analysis}}.}
\end{abstract}

\vspace{\vspacesize}
\section{Introduction}
\vspace{\vspacesize}

\sloppy

Given a set of $n$ samples of dimension $d$, denoted $\X =\left[\xbf_1, \cdots, \xbf_n\right] \in \R^{d\times n}$, a linear dimension reduction consists in projecting the data onto a $k$-dimensional subspace ($k<p$) as $\U^{\top} \X \in \R^{k\times n} $, where $\U \in {\rm St}(d,k)$ is an orthonormal basis (as $\St(d,k) = \{\U \in \R^{d \times k}, \U^{\top}\U = \mathbf{I}_k\}$ denotes the Stiefel manifold).
The most celebrated method in this framework is probably the principal component analysis (PCA) that selects the $k$ leading eigenvectors of $\mathbf{X}$ for the projection basis \cite{jolliffe2016principal}.

Interestingly, this basis appears as the solution to many underlying optimization problems, whose formulations offer points of view to generalize PCA and thus alleviate several of its shortcomings.
For example, PCA minimizes the average squared distance between the samples and their projection on the subspace spanned by $\U$.
Generalizations can then come from considering the minimization of robust distances to be less sensitive to outliers \cite{ding2006r, lerman2018fast}.
A second example is that any formulation of PCA as an optimization problem can be regularized to promote certain properties of the solution.
This is the starting point of many sparse PCA algorithms that aim to obtain a sparse basis $\U$, \ie, promoting the projection to act as a variable selection \cite{zou2006sparse}.

\begin{figure}[t]
    \centering
    \includegraphics[width = \linewidth]{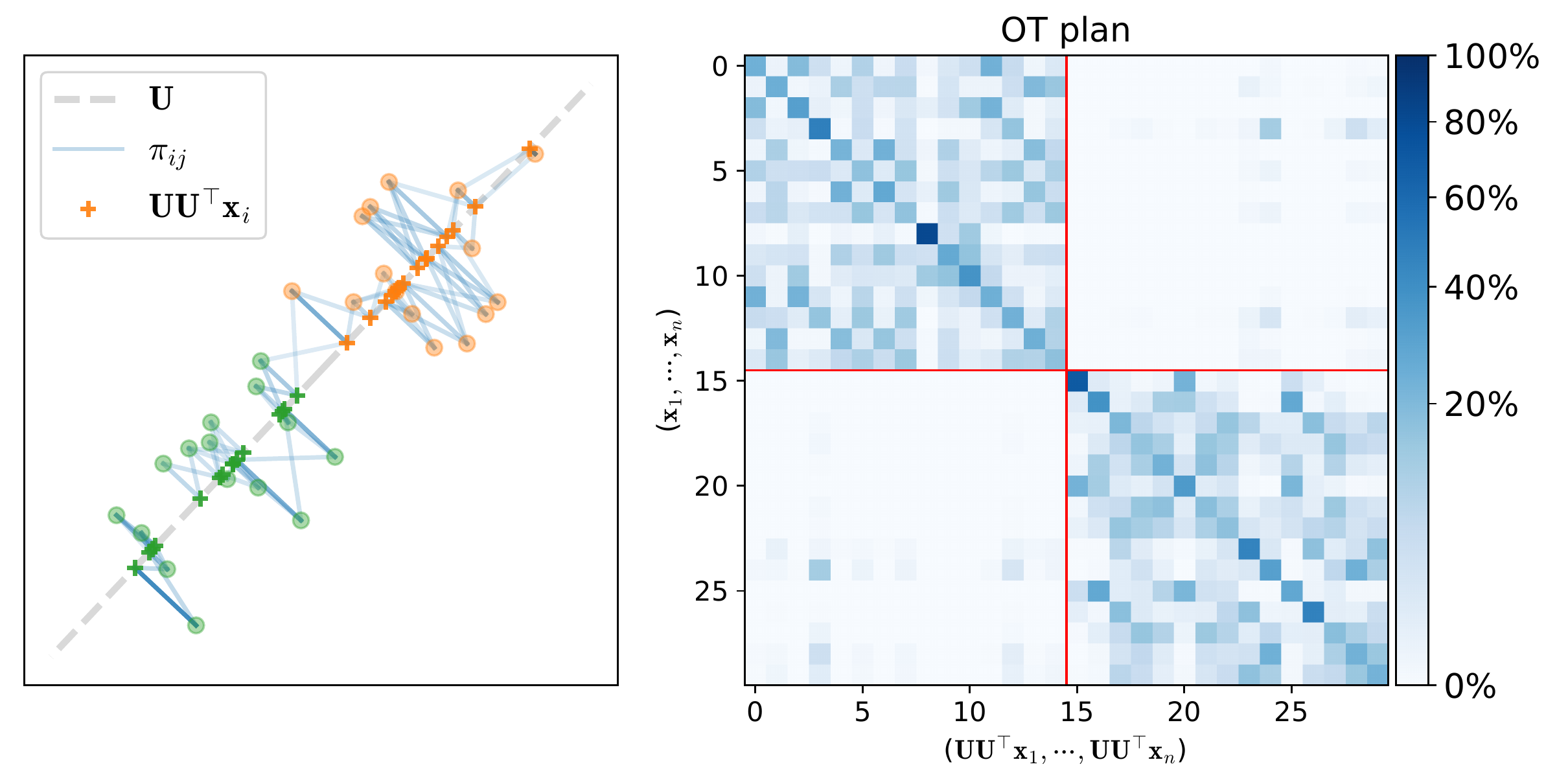}
    \caption{Illustration of \eqref{eq:ot_pca} with 2D samples organized in two clusters (in green and orange). On the left are the samples and their 1D projections, and on the right is the corresponding OT transport plan.}
    \label{fig:illustration_method}
\end{figure}

In this work, we explore a reformulation of PCA as the solution to an optimal transport (OT) problem \cite{P19}.
We show that optimizing an entropic OT between samples $\X$ and their projected counterparts $\U\U^{\top}\X$ encodes neighborhood information between samples. Interestingly, optimizing exact OT (in the special case in the absence of entropic regularization) is equivalent to standard PCA.
Thanks to the underlying principles of OT, this new approach is able to capture both global linear embeddings as well as local interactions between samples.

The contributions are the following:
$i$) We reformulate a subspace recovery problem with an OT objective and show that it indeed yields the standard PCA when using least-squares cost and no regularization;
$ii$) We propose a block-coordinate descent (BCD) algorithm to solve the corresponding optimization problem and a more efficient alternative using the majorization-minimization framework \cite{B21};
$iii$) We perform numerical experiments on genome data \cite{K01, F19}  that
show  a gain in accuracy compared to the standard PCA when used as a
preprocessing step for classification problems.

\vspace{\vspacesize}
\section{Entropic Wasserstein Component Analysis (EWCA)}
\vspace{\vspacesize}

\paragraph*{Entropic Optimal Transport.}
 Given two datasets $\X = (\xbf_1, \cdots , \xbf_n), \Y = (\ybf_1, \cdots,
 \ybf_m)$ with $\xbf_i, \ybf_j \in \R^{d}$ and $c: \R^{d} \times \R^{d}
 \rightarrow \R_{+}$ a cost function. Consider two histograms $\abf \in
 \Sigma_n, \bbf \in \Sigma_m$ (\ie\ $a_i \geq 0, \sum_{i=1}^{n} a_i = 1$) the entropic OT problem aims at solving, for $\varepsilon > 0$,
\begin{equation*}
    \OT_{\varepsilon,c}(\abf,\bbf,\X,\Y) \stackrel{\D}{=} \min_{\GG \in \Pi(\abf,\bbf)} \sum_{i,j=1}^{n,m} c(\xbf_i,\ybf_j)\pi_{ij} -\varepsilon \operatorname{H}(\GG) \,,
\end{equation*}
where $\Pi(\abf,\bbf) = \{\GG \in \R_{+}^{n \times m}; \ \GG \one_m = \abf,
\GG^{\top} \one_n = \bbf \}$ is the set of couplings between $\abf,\bbf$ and
$\operatorname{H}(\GG)=-\sum_{ij} \log(\pi_{ij}/a_i b_j)\pi_{ij}$ is the negative entropy. To simplify the notations we write $\OT_{\varepsilon,c}(\X,\Y)$ when $n=m$ and $\abf = \bbf =
\frac{1}{n}\one_n$. This problem can be solved using the 
Sinkhorn-Knopp (SK) algorithm~\cite{C13}. Specifically, given the Gibbs kernel $\mathbf{K} = \exp(-\mathbf{C}/\varepsilon)$, SK alternates (until convergence) the two steps
\begin{align*}
    \mathbf{u} & \leftarrow \mathbf{a} \oslash \mathbf{K}\mathbf{v}  \text{ // Update left scaling }          \\
    \mathbf{v} & \leftarrow \mathbf{b} \oslash \mathbf{K}^{\top} \mathbf{u}  \text{ // Update right scaling }
\end{align*}
and returns the coupling $\GG = \operatorname{diag}(\mathbf{u}) \mathbf{K}
\operatorname{diag}(\mathbf{v})$. SK involves simple iterations of matrix-vectors products that can run efficiently on GPU.


\paragraph*{Entropic Wasserstein Component Analysis.}
The principle of our method is to consider the optimization problem
\begin{equation}
    \label{eq:ot_pca}
    \tag{EWCA}
    \min_{\U \in \St(d,k)} \OT_{\varepsilon,c}(\X,\U \U^{\top} \X)\,,
\end{equation}
with the classical squared $\ell_2$ cost function $c(\xbf, \ybf) = \|\xbf - \ybf\|_2^2$.
Problem~\eqref{eq:ot_pca} is a non-convex problem that equivalently writes
\begin{equation}
    \label{eq:ot_pca_2}
    \min_{ \begin{smallmatrix} \GG \in \Pi(\frac{1}{n}\one_n,\frac{1}{n}\one_n) \\ \U \in \St(d,k) \end{smallmatrix}} \sum_{i,j =1}^{n,n}  \| \xbf_i - \U\U^{\top}\xbf_j \|_2^2 \ \pi_{ij} - \varepsilon \operatorname{H}(\GG).
\end{equation}
The objective function shares similarities with PCA but with key differences: the OT plan $\GG$ assigns weights between original and projected samples, while the entropic regularization adjusts the spread of mass between them. Therefore, as shown in Figure \ref{fig:illustration_method}, the OT plan weights the  samples within the neighborhood of the projected points, promoting the clustering of these points.  Note that a similar strategy was used in
Wasserstein Discriminant Analysis \cite{WDA} to find a discriminant subspace for
the data with a different objective.
\eqref{eq:ot_pca} is a difficult non-convex problem, and we propose two algorithms to solve it in the next section.

\paragraph*{Limit cases.}
Interstingly, when $\varepsilon \rightarrow 0$, we have $\GG \rightarrow
\frac{1}{n} \mathbf{I}_n$ and we recover the PCA objective.
Conversely when $\varepsilon \rightarrow +\infty, \ \GG \rightarrow
\frac{1}{n} \one_n \one_n^{\top}$. From the reformulation \eqref{eq:min_U_equivalent}
in the next section 
and when the data are centered (\ie\ $\X
\one_n = 0$), solving \eqref{eq:ot_pca_2} in $\U$ when $\varepsilon \rightarrow +\infty$ corresponds to  $\min_{\U
\in \St(d,k)} \tr\left(\U^{\top}[\frac{1}{n} \X \X^{\top}]\U\right)$ that is
finding the $k$ eigenvectors corresponding to the $k$ lowest eigenvalues of the
empirical covariance matrix. Therefore, $\varepsilon$ allows to interpolate between estimating eigenvectors of the empirical covariance matrix for the $k$ highest and lowest eigenvalues.


\begin{algorithm}[t]
    \caption{\label{alg:bcd} BCD for solving~\eqref{eq:ot_pca_2} when $\varepsilon > 0$}
    \begin{algorithmic}[1]
        \State $n_{it}, \mathbf{a}$, $\mathbf{b}$, $\epsilon >0$, $\U^{(0)}$
        \While{not converged}
        \State Let $\mathbf{C}^{(t)} = (c(\xbf_i, \U^{(t)}(\U^{(t)})^{\top}\xbf_j))_{ij}$
        \State Find $\GG^{(t)}$ with Sinkhorn-Knopp algorithm
        \State Let $\Mbf^{(t)} = \X \left(2 \sym(\GG^{(t)}) -\frac{1}{n} \mathbf{I}\right)\X^{\top}$
        \State Find $\ubf_1, \cdots, \ubf_k$ the eigenvectors corresponding the the $k$ highest eigenvalues of $\Mbf^{(t)}$
        \State $\U^{(t+1)} = (\ubf_1, \cdots, \ubf_k)$
        \State $t = t + 1$
        \EndWhile
        \State \Return $\U^{(t)}, \GG^{(t)}$
    \end{algorithmic}
\end{algorithm}

\vspace{\vspacesize}
\section{\label{sec:optim} Optimization algorithms for Entropic Wasserstein Component Analysis}
\vspace{\vspacesize}

\paragraph*{Block coordinate descent (BCD).}
A first approach to tackle the optimization problem~\eqref{eq:ot_pca} is presented with BCD algorithm.
Indeed, the cost function in~\eqref{eq:ot_pca_2} can be minimized by alternating a minimization over $\GG \in \Pi(\frac{1}{n}\one_n,\frac{1}{n}\one_n)$ (with SK algorithm) and a minimization over $\U \in \St(d,k)$ with fixed $\GG$. The latter step requires solving
\begin{equation}
    \label{eq:min_U}
    \min_{\U \in \St(d,k)} \sum_{i,j}^{n,n}  \| \xbf_i - \U\U^{\top}\xbf_j \|_2^2 \pi_{ij}\,.
\end{equation}
As described in Lemma \ref{lemma:bcd_u}, problem \eqref{eq:min_U} is minimized by finding the $k$ eigenvectors of $\Mbf \triangleq \X \left(2 \sym(\GG) - \frac{1}{n} \mathbf{I}\right)\X^{\top}$ associated with the $k$ highest eigenvalues.
Consequently, the BCD procedure, summarized in Algorithm~\ref{alg:bcd}, alternates between Sinkhorn-Knopp and computing the eigenvectors of $\Mbf$.
The complexity of the BCD is presented in Table~\ref{tab:complexities}.

\begin{lemma}
    \label{lemma:bcd_u}
    For any $\GG \in \Pi(\frac{1}{n}\one_n,\frac{1}{n}\one_n)$ the problem~\eqref{eq:min_U} is equivalent to
    \begin{equation}
        \label{eq:min_U_equivalent}
        \max_{\U \in \St(d,k)} \tr\left(\U^{\top} \Mbf \U\right)
    \end{equation}
    where
    \begin{equation}
        \label{eq:M_formula}
        \Mbf \triangleq \X \left( 2 \sym(\GG) - \frac{1}{n} \mathbf{I}\right)\X^{\top}
    \end{equation}
    with $\sym(\GG) \triangleq \left(\GG+ \GG^\top\right)/2$.
    Hence, the solution of \eqref{eq:min_U} is given by $\U^{\star}= (\ubf_1, \cdots, \ubf_k)$ where the $\ubf_i$'s are the eigenvectors corresponding the the $k$ highest eigenvalues of $\Mbf$.
\end{lemma}
\begin{proof}
    The cost can be written as $\frac{1}{n} \sum_{i} \|\xbf_i\|_2^{2}+\frac{1}{n} \sum_i \|\U \U^{\top}\xbf_i\|_2^{2} 
            -2 \sum_{ij} \langle \xbf_i, \U \U^{\top}\xbf_j \rangle \pi_{ij}$.
    The second term writes $\frac{1}{n} \sum_{i} \tr(\U \U^{\top} \xbf_i \xbf_i^{\top}) = \tr(\U \U^{\top} \frac{1}{n} \X \X^{\top})$ and the third term $-2 \tr(\U \U^{\top} \X \GG \X^{\top}) = -2 \tr(\U \U^{\top} \X \sym(\GG) \X^{\top})$ since $\X^{\top} \U \U^{\top} \X$ is symmetric.
    By combining both, we obtain \eqref{eq:min_U_equivalent}.
    To conclude, we used the Ky-Fan theorem.
\end{proof}

\paragraph*{Block-majorization-minimization (block-MM).} BCD is a simple approach but can become slow on high dimensional data due to the  computational complexity of $\Ocal(d^3)$.
To alleviate this problem, we seek to solve~\eqref{eq:min_U} without actually computing a $p \times p$ matrix by relying on block-MM algorithms over $\St(d,k)$~\cite{B21}.
In the next Lemma, we first formulate a problem that is equivalent to~\eqref{eq:min_U}
when restricted to $\St(d,k)$, but whose objective can be globally majorized on $\mathbb{R}^{d \times k}$ by a linear function.

\begin{lemma}
    \label{lemma:mm_u}
    For any $\GG \in \Pi(\frac{1}{n}\one_n,\frac{1}{n}\one_n)$ the problem~\eqref{eq:min_U} is equivalent to the following minimization problem
    \begin{equation}
        \label{eq:one_mm_u}
        \min_{\U \in \St(d,k)} \tr\left(\U^{\top} \Pbf \U \right)
    \end{equation}
    where $\Pbf \preccurlyeq \mathbf{0}$ is the matrix defined as
    \begin{multline}
        \label{eq:P_formula}
        \Pbf \triangleq \alpha_{\GG} \left(\Sigmab - 1_{\alpha_{\GG} > 0} \lambda_{\rm max}^{\Sigmab} \Ib\right) \\ - 2 \X \left(\sym(\GG) - \lambda_{\rm min}^{\sym(\GG)} \Ib \right) \X^{\top},
    \end{multline}
    and where $\lambda_{\rm max}^{\Sigmab}$ is the largest eigenvalue of $\Sigmab \triangleq \frac{1}{n} \X \X^{\top}$, $\lambda_{\rm min}^{\sym(\GG)}$ is the smallest eigenvalue of $\sym(\GG)$, $\alpha_{\GG} \triangleq 1 - 2n\lambda_{\rm min}^{\sym(\GG)}$, and $1_{\alpha_{\GG} > 0}$ is equal to $1$ if $\alpha_{\GG} > 0$ and $0$ otherwise.
\end{lemma}
\begin{proof}
    The problem~\eqref{eq:min_U_equivalent} is rewritten as
    \begin{equation*}
        \min_{\U \in \St(d,k)}  \tr\left(\U^{\top} \Sigmab  \U\right)\\
        - 2 \tr\left(\U^{\top} \X \sym(\GG) \X^{\top} \U\right)
    \end{equation*}
    which can also be rewritten as
    \begin{multline*}
        \min_{\U \in \St(d,k)}  \alpha_{\GG} \tr\left(\U^{\top} \Sigmab  \U\right)\\
        - 2 \tr\left(\U^{\top} \X \left(\sym(\GG) - \lambda_{\rm min}^{\sym(\GG)} \Ib \right) \X^{\top} \U\right).
    \end{multline*}
    Finally, we observe that when $\alpha_{\GG}$ is non-positive $\U \mapsto \alpha_{\GG} \tr\left(\U^{\top} \Sigmab  \U\right)$ is concave over $\R^{d\times k}$ since $\Sigmab \succcurlyeq \mathbf{0}$.
    \sloppy Otherwise (\ie \ $\alpha_{\GG}$ is positive), 
    we remark that the restriction of
    $\alpha_{\GG}\tr\left(\U^{\top} \Sigmab  \U\right)$
    to $\St(d,k)$ coincide with the concave function $\U \mapsto \alpha_{\GG}\tr\left(\U^{\top} \left(\Sigmab - \lambda_{\rm max}^{\Sigmab} \Ib\right)  \U\right) + k\lambda_{\rm max}^{\Sigmab}$.
\end{proof}

\noindent
Given the current iterate $\Pbf\preccurlyeq \mathbf{0}$, the objective of \eqref{eq:one_mm_u} is concave over $\R^{d\times k}$, so it can be majorized by its first order Taylor expansion at the point $\U^{(l)} \in \St(d,k)$:
\begin{equation}
    \tr\left(\U^{\top} \Pbf \U \right) \leq 2\tr\left(\U^{\top} \Pbf \U^{{(l)}}\right) + {\rm const}.
\end{equation}
The minimizer of the above upper bound on $\St(d,k)$ is the orthogonal projection of $- \Pbf \U^{(l)} \in \R^{d \times k}$ onto $\St(d,k)$, \ie
\begin{equation}
    \label{eq:min_lower_bound_U}
    \U^{(l+1)} = \text{pf}(-\Pbf \U^{(l)})
\end{equation}
where $\text{pf}(\Abf)$ is the orthogonal factor from the polar factorization of $\Abf \in \R^{d \times k}$.
Multiple iterations of~\eqref{eq:min_lower_bound_U} correspond to an MM algorithm that reaches a critical point of~\eqref{eq:min_U}~\cite{B21}.
Moreover, the cost function~\eqref{eq:min_U} is invariant to the action of $\St(k, k)$.
Consequently, any operator that yields the span of $-\Pbf \U^{(l)}$ (\eg, a QR decomposition) is a valid alternative to 
$\text{pf}(\cdot)$ in order to compute the update~\eqref{eq:min_lower_bound_U}.
An additional reduction in computational cost is realized by directly computing the product $\Pbf\U^{(l)}$ as
\begin{multline}
    \label{eq:PU_formula}
    \Pbf \U^{(l)} = \alpha_{\GG} \left[\frac{1}{n} \X \left( \X^{\top}\U^{(l)} \right) - 1_{\alpha_{\GG} > 0} \lambda_{\rm max}^{\Sigmab} \U^{(l)}\right] \\
    - 2 \X \left[\sym(\GG) - \lambda_{\rm min}^{\sym\left(\GG\right)} \Ib \right] \left( \X^{\top} \U^{(l)} \right).
\end{multline}
Hence, we transformed the BCD update that computes a $d \times d$ matrix and its SVD by computing a $d \times k$ matrix and its QR decomposition.
This strategy proves effective when $k \ll d$.
The overall block-MM procedure with the $\text{qf}$ function that returns the orthogonal factor of the QR decomposition and its complexity are summarized in Algorithm~\ref{alg:mm} and Table~\ref{tab:complexities} respectively.

\begin{algorithm}[t]
    \caption{\label{alg:mm} Block-MM for solving~\eqref{eq:ot_pca_2} when $\varepsilon > 0$}
    \begin{algorithmic}[1]
        \State $n_{it}, m_{it}, \mathbf{a}$, $\mathbf{b}$, $\epsilon >0$, $\U^{(0)}$
        \While{not converged}
        \State Let $\mathbf{C}^{(t)} = (c(\xbf_i, \U^{(t)}(\U^{(t)})^{\top}\xbf_j))_{ij}$
        \State Find $\GG^{(t)}$ with Sinkhorn-Knopp algorithm
        \For{ $ \U^{(l=0)} = \U^{(t)}$, $l=1,\dots,m_{it}$ }
        \State Compute $\Pbf^{(t)}  \U^{(l)}$ as in~\eqref{eq:PU_formula}
        \State $\U^{(l+1)} =  \text{qf}(\Pbf^{(t)} \U^{(l)})$ \quad \text{\# QR orth.}
        \EndFor
        \State $ \U^{(t)} = \U^{(l+1)}$
        \State $t = t+1$
        \EndWhile
        \State \Return $\U^{(t)}, \GG^{(t)}$
    \end{algorithmic}
\end{algorithm}

\begin{table*}[t]
    \centering
    \begin{tabular}{c|c|c}
        \hline
        \multirow{2}{*}{\shortstack{Common steps to BCD (Alg.~\ref{alg:bcd})                                                                         \\ and Block-MM (Alg.~\ref{alg:mm})}} & Computation of $\mathbf{C}^{(t)} = (c(\xbf_i, \U^{(t)}(\U^{(t)})^{\top}\xbf_j))_{ij}$  &  $\Ocal(n^2d)$ \\
                                                      & Computation of $\GG^{(t)}$                 & $\Ocal(n^2)$               \\
        \hline
        \multirow{3}{*}{BCD: scales with $n$ (Alg.~\ref{alg:bcd})}     & Computation of $\Mbf^{(t)}$ (Eq.~\eqref{eq:M_formula})          & $\Ocal(n^2d + nd^2)$       \\
                                                      & Computation of the eigenvectors of $\Mbf$                       & $\Ocal(d^3)$               \\
                                                      & Overall complexity                                              & $\Ocal(n^2d + nd^2 + d^3)$ \\
        \hline
        \multirow{3}{*}{Block-MM: scales with $d$ (Alg.~\ref{alg:mm})} & Computation of $\Pbf^{(t)}\U^{(l)}$ (Eq.~\eqref{eq:PU_formula}) & $\Ocal(ndk + n^3)$         \\
                                                      & Projection of $\Pbf^{(t)}\U^{(l)}$ onto $\St(d,k)$              & $\Ocal(dk^2)$              \\
                                                      & Overall complexity                                              & $\Ocal(n^2d + n^3)$        \\
        \hline
    \end{tabular}
    \caption{Comparison of the computation complexities of Algorithms~\ref{alg:bcd} and~\ref{alg:mm} with respect to the executed steps.}
    \label{tab:complexities}
\end{table*}

\paragraph*{Computation of $\lambda_{\rm max}^{\Sigmab}$ and $\lambda_{\rm min}^{\sym(\GG)}$.}
Finally, we remark that the computation of the eigenvalues $\lambda_{\rm max}^{\Sigmab}$ and $\lambda_{\rm min}^{\sym(\GG)}$ is not a limitation.
Indeed, in practice, both Algorithms~\ref{alg:bcd} and~\ref{alg:mm} are initialized with the PCA.
Thus, the $\lambda_{\rm max}^{\Sigmab}$ is obtained from this initialization.
Then, $\lambda_{\rm min}^{\sym(\GG)}$ can be replaced by any lower bound.
Since $\sym(\GG) \in \Pi(\frac{1}{n}, \frac{1}{n})$, and using the Gershgorin circle theorem, a lower bound of $\lambda_{\rm min}^{\sym(\GG)}$ is $-\frac{1}{n}$.
This replacement is convenient when $n$ is large since it avoids the computation of the SVD of $\sym(\GG)$.
As the corresponding majorizer is less tight, we observed in practice a slightly lower rate of convergence in iterations when replacing $\lambda_{\rm min}^{\sym(\GG)}$ by $-\frac{1}{n}$ in Equation~\eqref{eq:PU_formula}.

\begin{figure*}[t]
    \centering
    \begin{subfigure}[b]{0.5\textwidth}
         \centering
         \includegraphics[height=\sizefig]{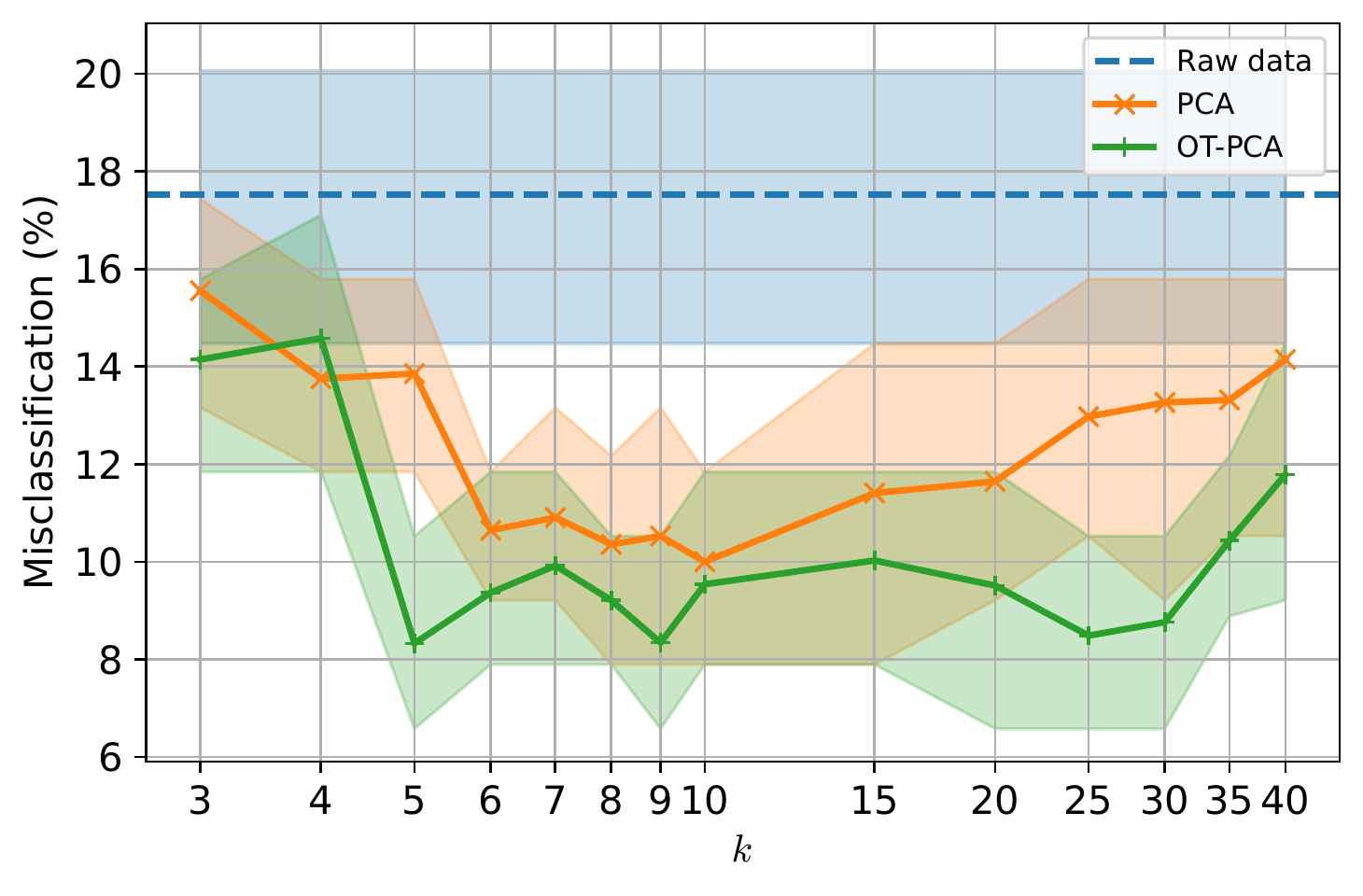}
         \caption{\emph{Breast} dataset}
         \label{fig:Breast_mis_k}
     \end{subfigure}\hfill
     \begin{subfigure}[b]{0.5\textwidth}
         \centering
         \includegraphics[height=\sizefig]{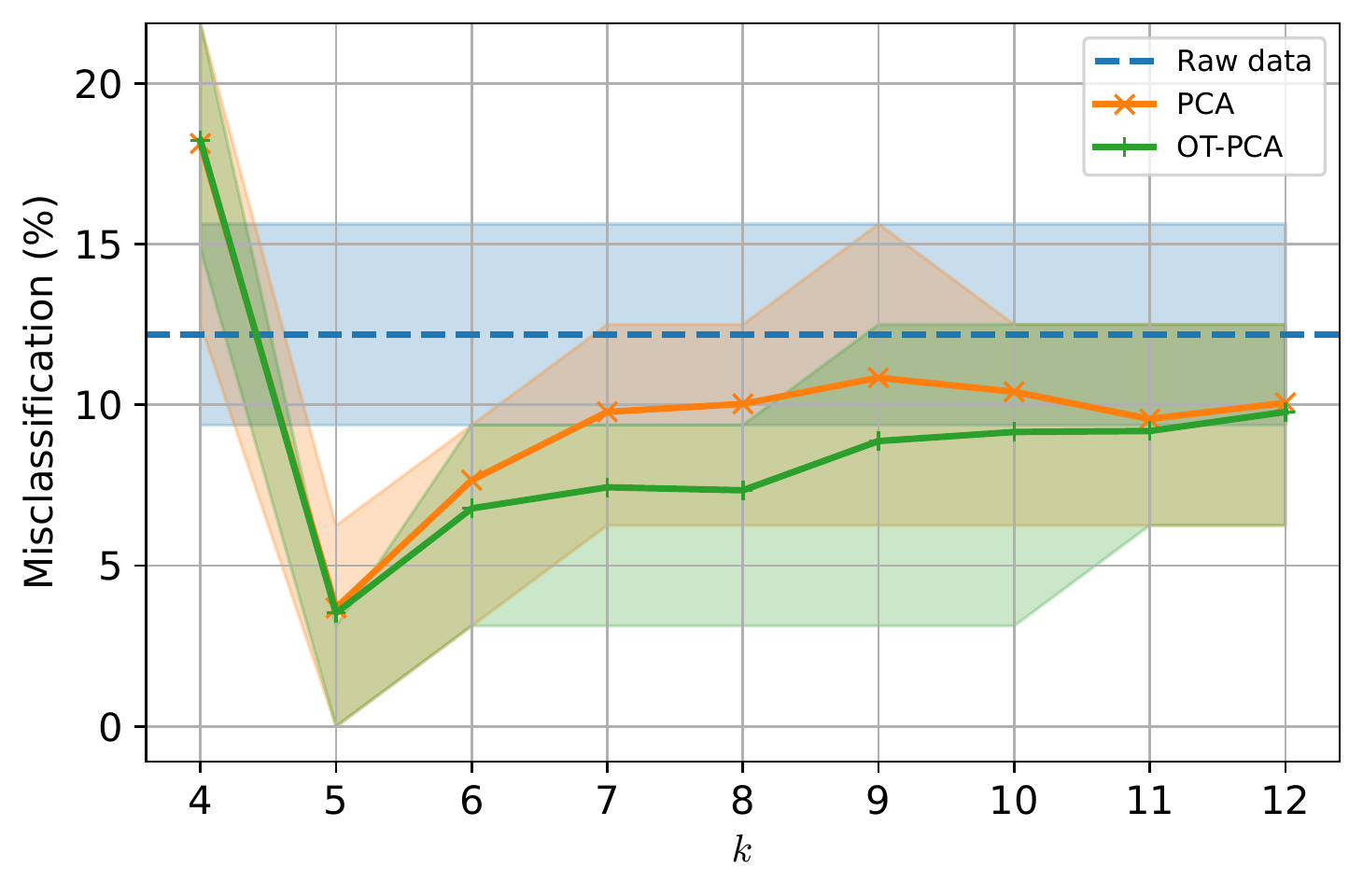}
         \caption{\emph{khan2001} dataset}
         \label{fig:Khan2001_mis_k}
     \end{subfigure}
    \caption{
        \textbf{Misclassification rate ($\%$) versus subspace dimension $k$ (the lower the better)}.
        Data are classified using a $1$ nearest neighbor classifier on $100$ splits train-test ($50\%-50\%$).
        In addition to the raw data (no preprocessing), two preprocessing are considered: PCA and EWCA (proposed in Algorithm~\ref{alg:mm}).
        The value of $\varepsilon$ of EWCA is chosen as the best performing on $20$ splits of the train set.
        The mean misclassification and the $1^\text{st}$ and $3^\text{rd}$ quartiles are reported.
    }
    \label{fig:mis_k}
\end{figure*}

\vspace{\vspacesize}
\section{Numerical experiments}
\vspace{\vspacesize}

\begin{figure*}[t]
    \centering
    \begin{subfigure}[b]{0.5\textwidth}
         \centering
         \includegraphics[height=\sizefig]{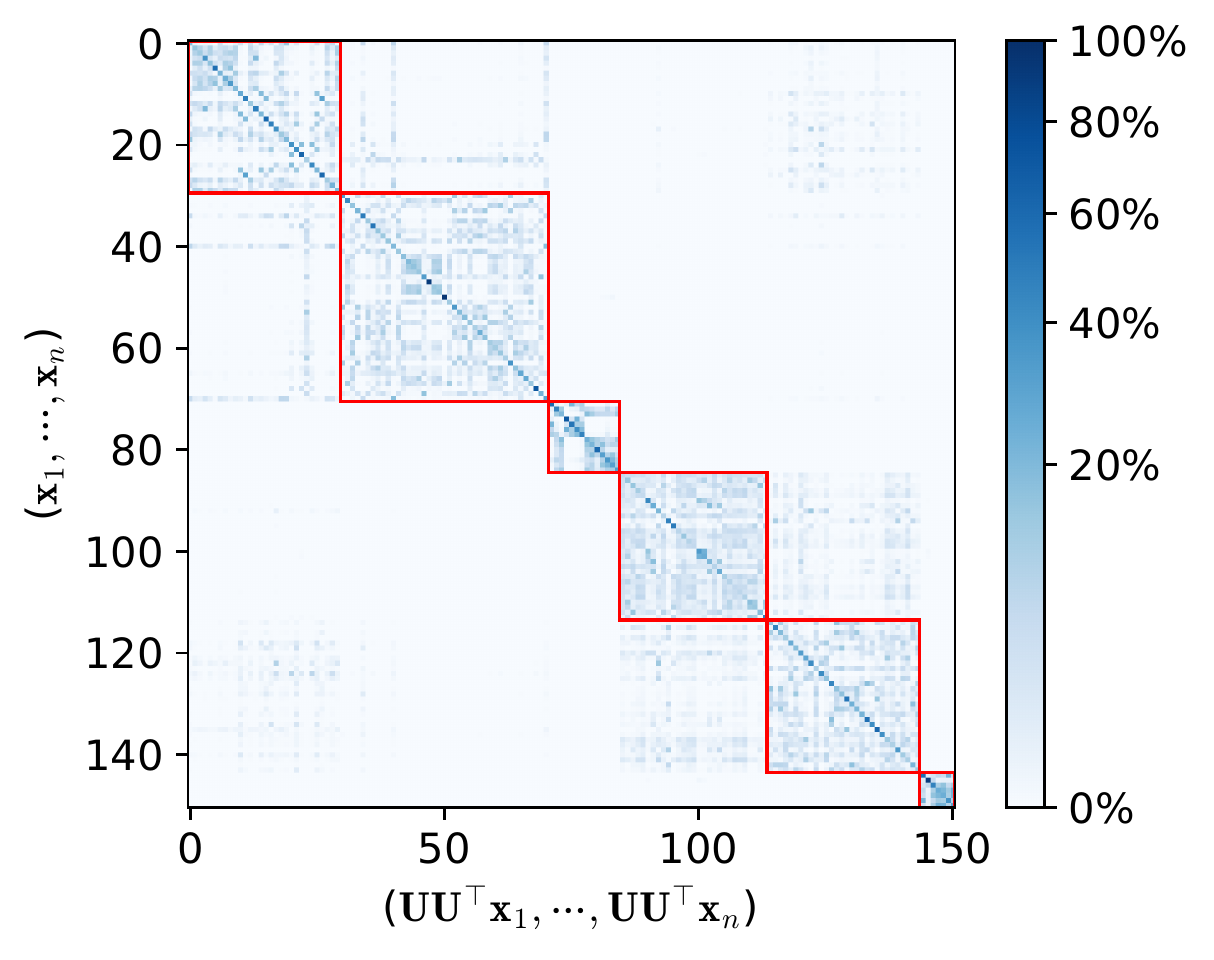}
         \caption{\emph{Breast} dataset}
        \label{fig:Breast_transp_plan}
     \end{subfigure}\hfill
     \begin{subfigure}[b]{0.5\textwidth}
         \centering
         \includegraphics[height=\sizefig]{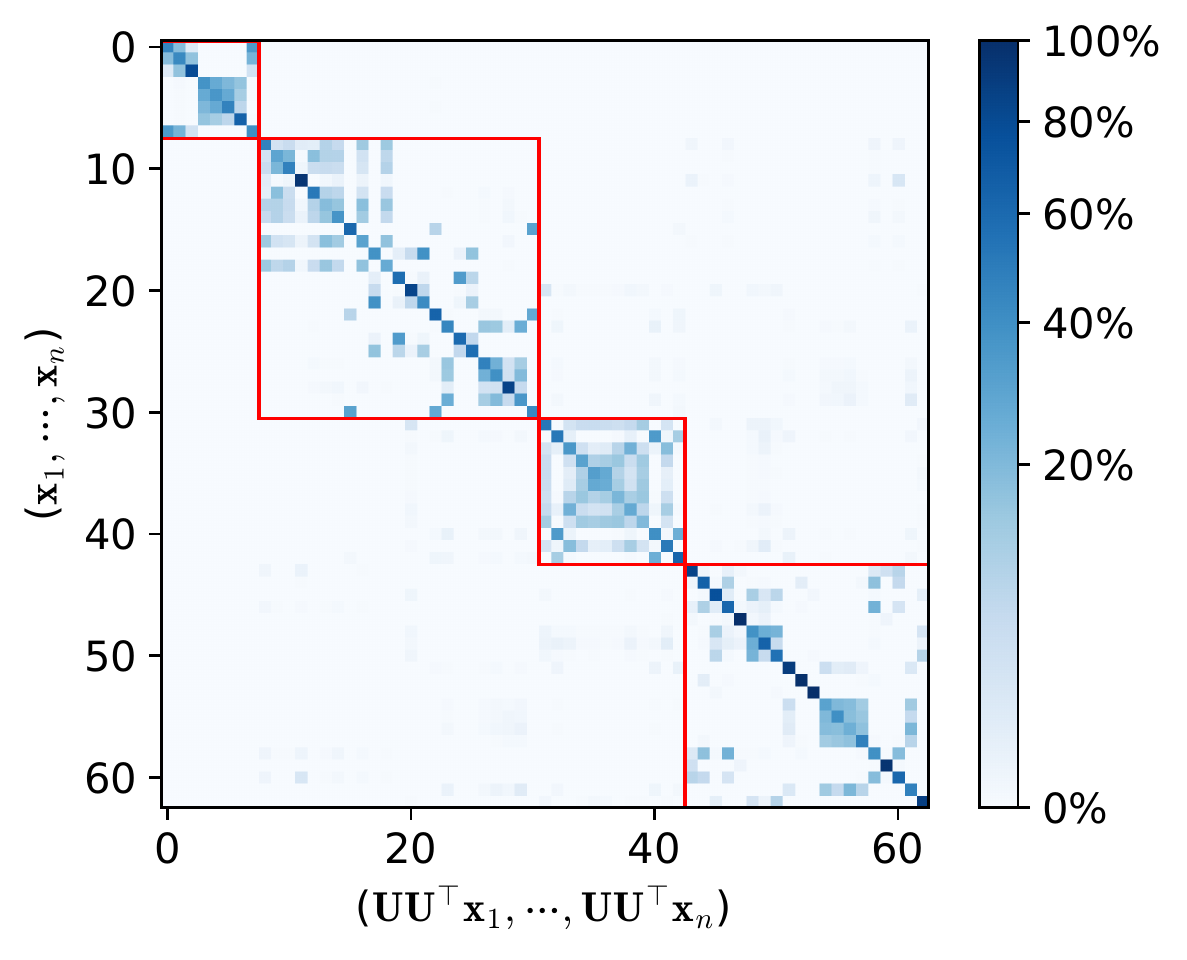}
         \caption{\emph{khan2001} dataset}
        \label{fig:Khan2001_transp_plan}
     \end{subfigure}
    \caption{
        \textbf{Transport plan $\GG$ ($\%$) computed on the \emph{Breast} dataset with EWCA (Alg.~\ref{alg:mm}) between raw data $(\xbf_1, \cdots, \xbf_n)$ and their projected counterparts $(\U\U^{\top}\xbf_1, \cdots, \U\U^{\top}\xbf_n)$}.
        $k=5$ and the values of $\varepsilon$ are the one performing the best on Figure~\ref{fig:mis_k}.
        The red squares enclose the data belonging to the same class.
        Classes are ordered from left to right and top to bottom, \ie the red square on the top left corner is class 1, and the red square on the bottom right corner is the last class.
     }
     \label{fig:transp_plan}
\end{figure*}

To assess the performance of the developed Algorithms~\ref{alg:bcd} and~\ref{alg:mm}, we leverage two classification datasets: \emph{Breast}~\cite{F19} and \emph{khan2001} \cite{K01}.
The \emph{Breast} dataset contains $n=151$ samples with $d=54675$ gene expressions each.
The goal is to classify these data into $6$ classes corresponding to breast cancer subtypes and normal tissues.
The \emph{Khan2001} dataset contains $n=63$ samples of $d=2308$ gene expression profiles to classify into $4$ types of tumors of childhood.
Notice that the interpolation effect induced by $\epsilon$ in EWCA
trades some explained variance (maximized by the PCA)
for an alternate representation of the data.
The relevance and quality of this representation are analyzed through quantitative and qualitative experiments on both datasets.
Moreover, the speed of the block-MM algorithm (Alg.~\ref{alg:mm}) over the BCD algorithm (Alg.~\ref{alg:bcd}) is shown in an experiment on the \emph{Breast} dataset.

\begin{figure}[t]
    \centering
    \includegraphics[height=\sizefig]{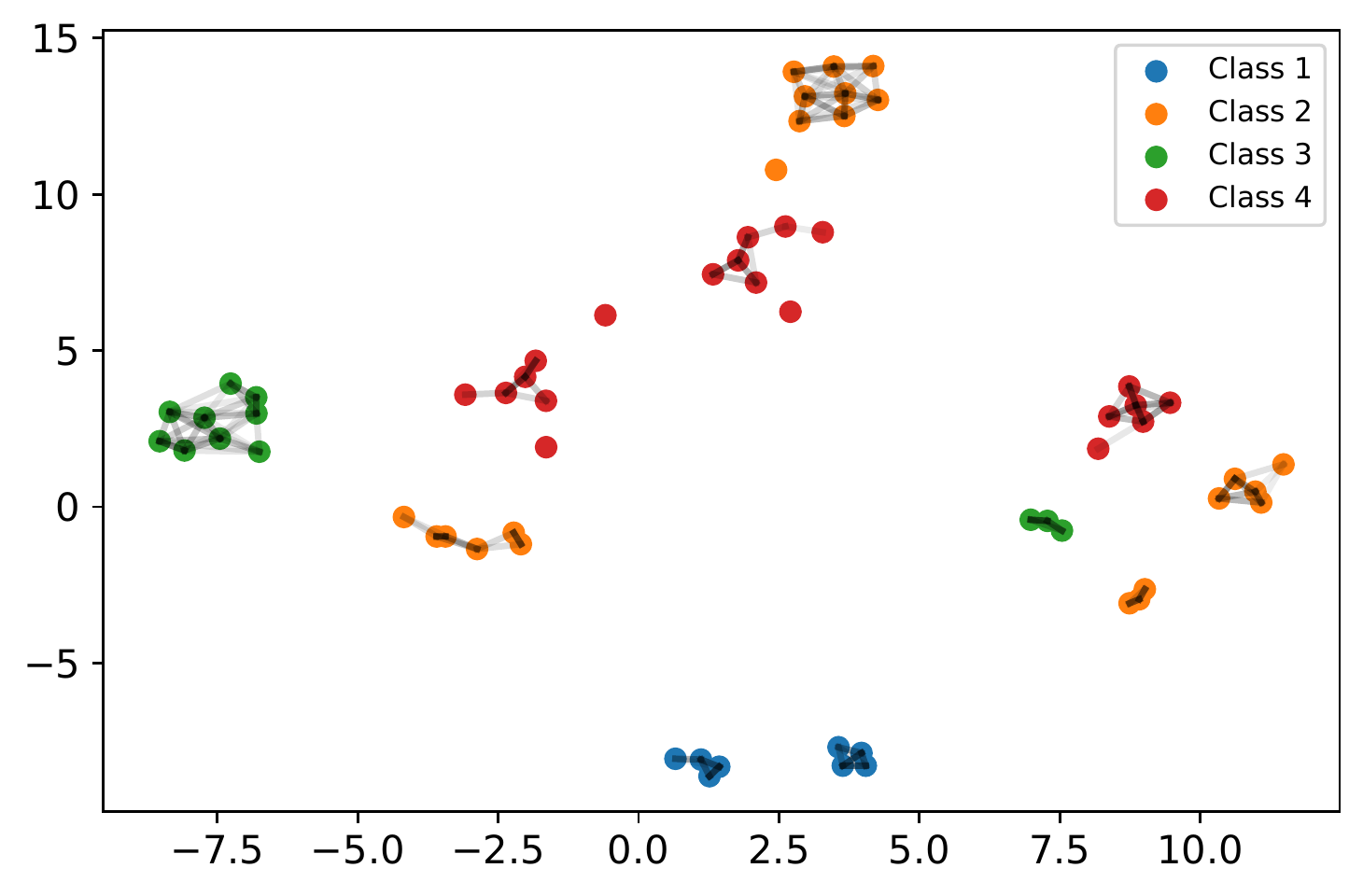}
    \caption{
        \textbf{TSNE of the projected data $(\U^{\top}\xbf_1, \cdots, \U^{\top}\xbf_n)$ computed with EWCA (Alg.~\ref{alg:mm}) on the \emph{Khan2001} dataset}.
        The subspace dimension is chosen as $k=5$ and the used subspace corresponds to the entropy intensity $\varepsilon$ performing the best on Figure~\ref{fig:Khan2001_mis_k}.
        The grey links represent the intensity of the values of the transport plan presented in Figure~\ref{fig:Khan2001_transp_plan} (values under a certain threshold are set to $0$).
    }
    \label{fig:Khan2001_TSNE}
\end{figure}

\begin{figure}[t]
    \centering
    \includegraphics[height=\sizefig]{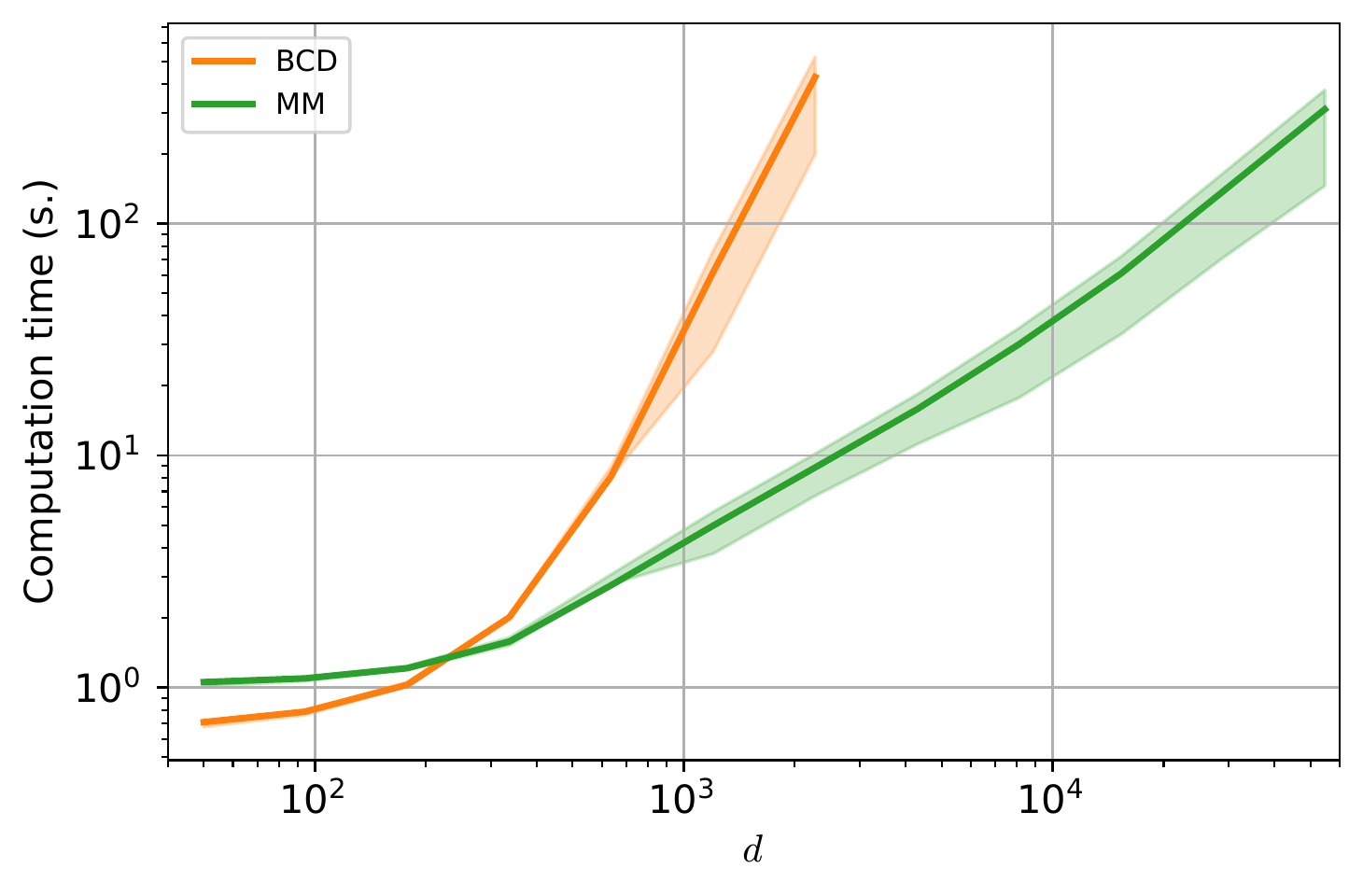}
    \caption{
        \textbf{Computation time in seconds of EWCA (Alg.~\ref{alg:mm}) for $k=5$ versus data dimension $d$ of subsampled genes of the \emph{Breast} dataset (the lower the better)}.
        The mean, $1^\text{st}$ and $3^\text{rd}$ quartiles computed with $100$ sets of subsampled genes are reported.
    }
    \label{fig:Breast_time}
\end{figure}

\begin{figure}[t]
    \centering
    \includegraphics[height=\sizefig]{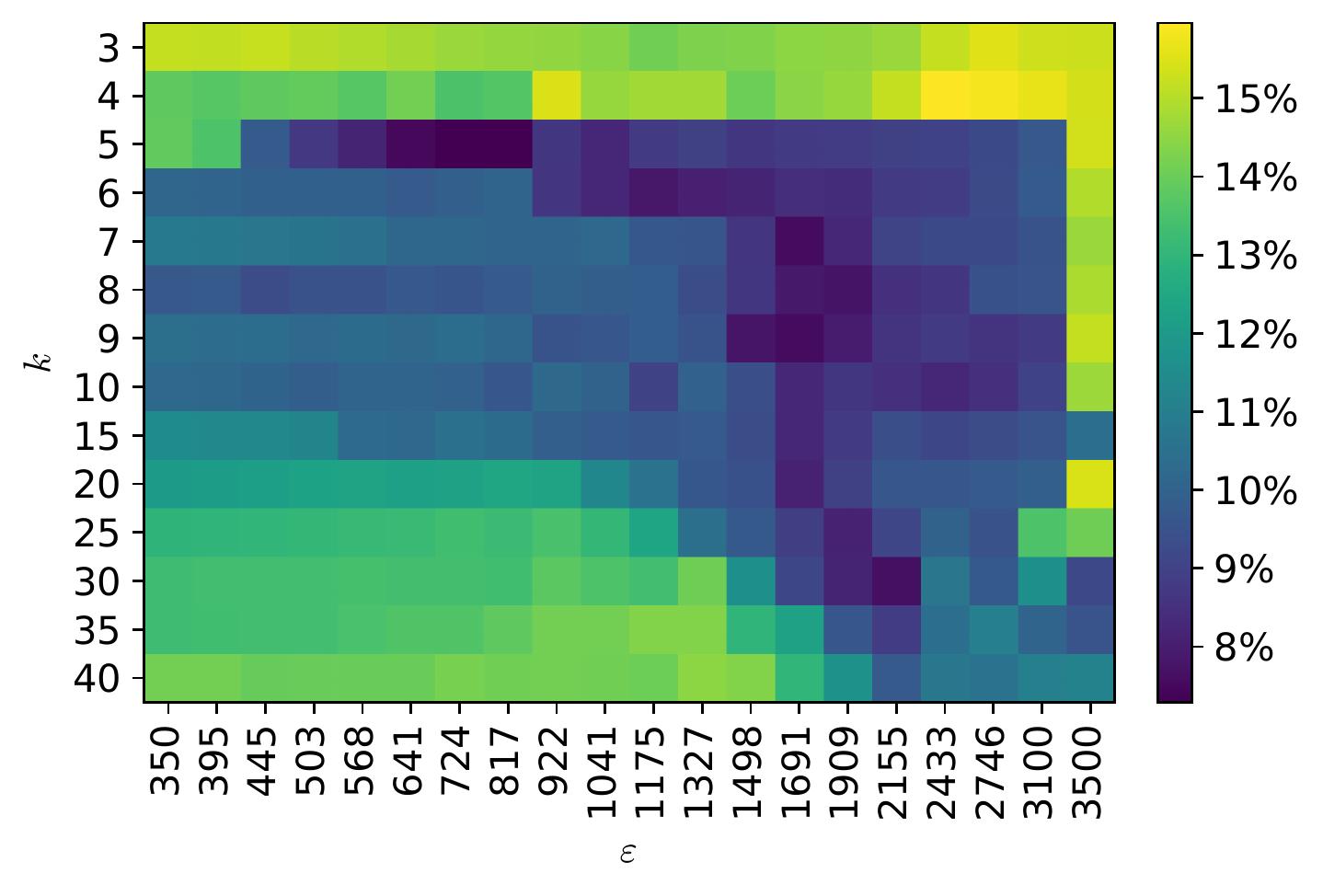}
    \caption{
        \textbf{Misclassification rate ($\%$) versus subspace dimension $k$ and entropy intensity $\varepsilon$ on the \emph{Breast} dataset (the lower the better)}.
        Data projected on the subspaces estimated by EWCA (proposed in Algorithm~\ref{alg:mm}) are classified using a $1$ nearest neighbor classifier on $100$ splits train-test ($50\%-50\%$).
    }
    \label{fig:Breast_mis_k_reg}
\end{figure}

\paragraph*{Classification performance.}
We first compute the misclassification rate, over $100$ train-test splits ($50\%-50\%$), of a $1$ nearest neighbor classifier on the raw data $(\xbf_1, \cdots, \xbf_n)$.
Then, we estimate subspaces with PCA and EWCA across their respective hyperparameters, \ie the subspace dimension $k$ and the intensity of the entropy regularization $\varepsilon$.
For each value of $k$, we compute the misclassification rates, over $100$ train-test splits ($50\%-50\%$), of a $1$ nearest neighbor classifier applied on the projected data $(\U^{\top}\xbf_1, \cdots, \U^{\top}\xbf_n)$.
The hyperparameter $\varepsilon$ of EWCA is chosen over $20$ splits on the train set.
The mean misclassification rates, as well as the $1^\text{st}$ and $3^\text{rd}$ quartiles, are reported across the different tested subspace dimensions $k$ in the Figure~\ref{fig:mis_k}.
We observe that EWCA and PCA give better accuracies than the $1$ nearest neighbor classifier applied directly on raw data, showing interest in considering dimension reduction methods.
Then, EWCA outperforms PCA on a wide range of values of $k$ on both datasets.
For certain values of $k$, the improvement in classification performance is large.
Indeed, on the \emph{Breast} dataset, at $k=5$, the misclassification rate is down from $14\%$ to $8\%$; \ie~a reduction of half of the error.
On the \emph{Khan2001} dataset, at $k=8$, the misclassification rate is down from $11\%$ to $7.5\%$; \ie~a reduction of a third of the error.
This improvement in discriminative capabilities indicates that EWPCA provides linear embeddings that favor clusters within samples in an unsupervised way.


\paragraph*{Transport plan interpretation.}

Then, Figure~\ref{fig:transp_plan} displays the transport plans $\GG$ estimated by EWCA at $k=5$ with $\varepsilon$ chosen as the best performing on the $20$ splits of the train set.
The data are ordered by class, and those that belong to the same class are enclosed in a red box.
We observe on both datasets that the transport plans values $\pi_{ij}$ are higher within data that belong to the same class than within data that belong to different classes.
This means that given a point $\xbf_i$ that belongs to the class $y_i$, the estimated subspace minimizes the discrepancy between $\xbf_i$ and the projected points $\U\U^{\top}\xbf_j$ that belong to the class $y_i$.
This enforces, in an unsupervised way, that points that belong to the same class are close to each other once projected in the estimated subspace.
Furthermore, using the transport plan from the \emph{Khan2001} dataset, several clusters can be identified.
Indeed, in the red square on the top left corner (class 1) of Figure~\ref{fig:Khan2001_transp_plan}, two clusters are distinguishable.
These two clusters are also observable in the samples from class 1 in Figure~\ref{fig:Khan2001_TSNE}.
The latter plots a TSNE~\cite{VDM08} of projected data $(\U^{\top}\xbf_1, \cdots, \U^{\top}\xbf_n)$ and the transport plan values.
This again indicates that EWCA identifies clusters by jointly estimating the transport plan and the subspace to project data on.

\paragraph*{Computation cost: block-MM versus BCD.}

So far, we have shown the good performance of EWCA in terms of precisions and given an interpretation of the estimated transport plan.
We now leverage the \emph{Breast} dataset to analyze the computational time of the proposed Algorithms~\ref{alg:bcd} and~\ref{alg:mm}.
Indeed, we subsample $d \in \llbracket 500, 54675 \rrbracket$ genes and run the two algorithms until convergence.
The mean computation time in seconds and the $1^\text{st}$ and $3^\text{rd}$ quartiles are reported.
When $d \leq 2000$, the BCD is faster than the block-MM thanks to its closed form formula on the $\U$-step.
However, when $d > 2000$, the block-MM algorithm is much faster than the BCD with a much lower rate of increase.
This illustrates the lower computation complexity in $d$ of the block-MM compared to the BCD one (see Table~\ref{tab:complexities}).

\paragraph*{Sensitivity to hyperparameter $\varepsilon$.}

In the classification tasks, we selected the hyperparameter $\varepsilon$ as the best performing one for a $1$ nearest neighbor classifier on $20$ splits of the train set.
To mitigate this necessity of testing many values of $\varepsilon$, we plot in Figure~\ref{fig:Breast_mis_k_reg} the heat map of the misclassification rates with respect to $k$ and $\varepsilon$ using the same protocol as the one used in Figure~\ref{fig:mis_k}.
On a wide range of $\varepsilon$, we observe that EWCA has similar misclassification rates as PCA, if not better.
Hence, EWCA is not too sensitive to the choice of $\varepsilon$.

\vspace{\vspacesize}
\section{Conclusion}
\vspace{\vspacesize}

We reformulated the PCA algorithm as the minimizer of the squared $2$-Wasserstein distance between a dataset and its projected counterpart.
Adding an entropy regularizer enabled us to consider pairs of points $(\xbf_i, \U\U^{\top}\xbf_j)$, with $i \neq j$ in this new optimization problem called EWCA.
To solve it, we proposed two algorithms, a BCD and a block-MM.
The latter showed faster convergence in high-dimensional regimes.
When leveraged as a preprocessing step for classification problems on gene expression datasets, we showed that EWCA yields a projection that favors clusters within the data in an unsupervised way.
The joint use of EWCA and its achieved transport map thus offers an interesting alternative to PCA for exploratory data analysis.


\paragraph*{Acknowledgements.}
Numerical experiments have realized with:
Matplotlib~\cite{matplotlib},
Scikit-learn~\cite{scikit-learn},
Numpy~\cite{numpy}, and POT~\cite{pot}.
\iftoggle{anonymous}{}{This work was supported by ANR MASSILIA (ANR-21-CE23-0038-01).}



\vspace{\vspacesize}
\bibliographystyle{IEEEbib}
\footnotesize
\bibliography{biblio}


\end{document}